\documentclass[11pt,a4paper]{article}
\usepackage[hyperref]{emnlp2018}
\usepackage{times}
\usepackage{latexsym}
\usepackage{tabularx}
\usepackage{graphicx}
\usepackage{array}
\usepackage{multirow}
\usepackage{amsmath}
\usepackage{booktabs}
\usepackage[toc,page]{appendix}
\usepackage{float}
\usepackage{makecell}

\usepackage{url}

\newcommand*\samethanks[1][\value{footnote}]{\footnotemark[#1]}

\aclfinalcopy 
\def\aclpaperid{1472} 

\title{Dissecting Contextual Word Embeddings: \\Architecture and Representation
}

\author{\makecell{Matthew E.~Peters\thanks{~~These authors contributed equally to this work.}$~~^1$, Mark Neumann\samethanks$~~^1$, Luke Zettlemoyer$^2$, Wen-tau Yih$^1$} \\
$^1$Allen Institute for Artificial Intelligence, Seattle, WA, USA\\
$^2$Paul G. Allen Computer Science \& Engineering, University of Washington\\
{\tt $\{$matthewp,markn,scottyih$\}$@allenai.org lsz@cs.washington.edu}\\}

\date{}

\begin{document}
\maketitle
\begin{abstract}
Contextual word representations derived from pre-trained bidirectional language models (biLMs) have recently been shown to provide significant improvements to the state of the art for a wide range of NLP tasks. However, many questions remain as to how and why these models are so effective. In this paper, we present a detailed empirical study of how the choice of neural architecture (e.g. LSTM, CNN, or self attention) influences both end task accuracy and qualitative properties of the representations that are learned. We show there is a tradeoff between speed and accuracy, but all architectures learn high quality contextual representations that outperform word embeddings for four challenging NLP tasks. Additionally, all architectures learn representations that vary with network depth, from exclusively morphological based at the word embedding layer through local syntax based in the lower contextual layers to longer range semantics such coreference at the upper layers. Together, these results suggest that unsupervised biLMs, independent of architecture, are learning much more about the structure of language than previously appreciated.
\end{abstract}

\section{Introduction}
Language model pre-training has been shown to substantially improve performance for many NLP tasks including question answering, coreference resolution, and semantic role labeling \cite{Peters:2018},
text classification \cite{Dai2015SemisupervisedSL,Howard:2018}, sequence tagging \cite{Peters2017SemisupervisedST}, sequence--to--sequence learning \cite{Ramachandran2017ImproveSeq2SeqLMGal2016ATG}, and constituency parsing \cite{Kitaev2018TransformerParsing,Joshi:2018}.
Despite large gains (typical relative error reductions range from 10--25\%), we do not yet fully understand why or how pre-training works in practice. In this paper, we take a step towards such understanding by empirically studying how the choice of neural architecture (e.g. LSTM, CNN, or self attention) influences both direct end-task accuracies and how contextualized word representations encode notions of syntax and semantics.



Previous work on learning contextual representations has used LSTM-based biLMs, but there is no prior reason to believe this is the best  possible architecture.
More computationally efficient networks have been introduced for sequence modeling including including gated CNNs for language modeling \cite{Dauphin2017LanguageMW} and feed forward self-attention based approaches for machine translation \cite[Transformer; ][]{Vaswani2017AttentionIA}.
As RNNs are forced to compress the entire history into a hidden state vector before making predictions while CNNs with a large receptive field and the Transformer may directly reference previous tokens, each architecture will represent information in a different manner. 

Given such differences, we study whether more efficient architectures can also be used to learn high quality contextual vectors.
We show empirically that all three approaches provide large improvements over traditional word vectors when used in state-of-the-art models across four benchmark NLP tasks.
We do see the expected tradeoff between speed and accuracy between LSTMs and the other alternatives, but the effect is relatively modest and all three networks work well in practice. 


Given this result, it is important to better understand what the different networks learn. In a detailed quantitative evaluation, we probe the learned representations and show that, in every case, they represent a rich hierarchy of contextual information throughout the layers of the network in an analogous manner to how deep CNNs trained for image classification learn a hierarchy of image features \cite{zeiler2014visualizing}.
For example, we show that in contrast to traditional word vectors which encode some semantic information, the word embedding layer of deep biLMs focuses exclusively on word morphology.
Moving upward in the network, the lowest contextual layers of biLMs focus on local syntax, while the upper layers can be used to induce more semantic content such as within-sentence pronominal coreferent clusters.
We also show that the biLM activations can be used to form phrase representations useful for syntactic tasks. Together, these results suggest that large scale biLMs, independent of architecture, are learning much more about the structure of language than previous appreciated. 




\section{Contextual word representations from biLMs}
To learn contextual word representations, we follow previous work by first training a biLM on a large text corpus (Sec.\ \ref{sec:bilms}).
Then, the internal layer activations from the biLM are transferred to downstream tasks (Sec.\ \ref{sec:elmo_desc}).


\subsection{Bidirectional language models}
\label{sec:bilms}
Given a sequence of $N$ tokens, $(t_1, t_2, ..., t_N)$, a biLM combines a forward and backward language model to jointly maximize the log likelihood of both directions:
\[
\begin{split}
\sum_{k=1}^N \left( \right. & \log p({t_k} \mid t_1, \ldots, t_{k-1}; \overrightarrow{\Theta}) \\
+  & \log p({t_k} \mid t_{k+1}, \ldots, t_{N}; \overleftarrow{\Theta})
\left. \right),
\end{split}
\]
where $\overrightarrow{\Theta}$ and $\overleftarrow{\Theta}$ are the parameters of the forward and backward LMs respectively.

To compute the probability of the next token, state-of-the-art neural LMs first produce a context-insensitive token representation or word embedding, $\mathbf{x}_k$, (with either an embedding lookup or in our case a character aware encoder, see below).
Then, they compute $L$ layers of context-dependent representations $\overrightarrow{\mathbf{h}}_{k,i}$ where $i \in [1, L]$ using a RNN, CNN or feed forward network (see Sec.\ \ref{sec:architectures}).
The top layer output $\overrightarrow{\mathbf{h}}_{k,L}$ is used to predict the next token using a Softmax layer.
The backward LM operates in an analogous manner to the forward LM.
Finally, we can concatenate the forward and backward states to form $L$ layers of contextual representations, or context vectors, at each token position: $\mathbf{h}_{k,i} = [\overrightarrow{\mathbf{h}}_{k,i}; \overleftarrow{\mathbf{h}}_{k,i}]$.
When training, we tie the weights of the word embedding layers and Softmax in each direction but maintain separate weights for the contextual layers.

\subsection{Character based language models}
\label{sec:character_models}
Fully character aware models \cite{kim2015characterNeuralLM} are considerably more parameter efficient then word based models but more computationally expensive then word embedding based methods when training.
During inference, these differences can be largely eliminated by pre-computing embeddings for a large vocabulary and only falling back to the full character based method for rare words.
Overall, for a large English language news benchmark, character aware models have slightly better perplexities then word based ones, although the differences tend to be small \cite{Jzefowicz2016ExploringTL}.

Similar to \citet{kim2015characterNeuralLM}, our character-to-word encoder is a five-layer sub-module that first embeds single characters with an embedding layer then passes them through 2048 character n-gram CNN filters with max pooling, two highway layers \cite{Srivastava2015TrainingVD}, and a linear projection down to the model dimension.

\subsection{Deep contextual word representations}
\label{sec:elmo_desc}
After pre-training on a large data set, the internal representations from the biLM can be transferred to a downstream model of interest as contextual word representations.
To effectively use all of the biLM layers, \citet{Peters:2018} introduced ELMo word representations, whereby all of the layers are combined with a weighted average pooling operation, \textbf{ELMo}$_k = \gamma \sum_{j=0}^L s_j \mathbf{h}_{k, j}$.
The parameters $\mathbf{s}$ are optimized as part of the task model so that it may preferentially mix different types of contextual information represented in different layers of the biLM.
In Sec.\ \ref{sec:elmo_eval} we evaluate the relative effectiveness of ELMo representations from three different biLM architectures vs.\ pre-trained word vectors in four different state-of-the-art models.

\begin{table*}[!t]
\centering
\begin{tabular}{l|c|c|c|c|c} \hline
\rule{0pt}{4ex} \textbf{Architecture} & \textbf{\# layers} & \textbf{Perplexity} & \textbf{\# params. (M)} & \parbox{14ex}{\textbf{Inference (ms) \\ 1 sentence}} & \parbox{14ex}{\textbf{Inference (ms) \\ 64 sentences}} \\[10pt] \hline\hline
LSTM & 2 & 39.7 & 76 / 94 & 44 / 46 & 66 / 85 \\
LSTM & 4 & 37.5 & 151 / 153 & 85 / 86 & 102 / 118 \\
Transformer & 6 & 40.7 & 38 / 56 & 12 / 13 & 22 / 44 \\
Gated CNN & 16 & 44.5 & 67 / 85 & 9 / 11 & 29 / 55 \\ \hline
\end{tabular}
\caption{\label{tab:table1}
Characteristics of the different biLMs in this study.  For each model, the table shows the number of layers used for the contextual representations, the averaged forward and backward perplexities on the 1 Billion Word Benchmark, the number of parameters (in millions, excluding softmax) and the inference speed (in milliseconds with a Titan X GPU, for sentences with 20 tokens, excluding softmax).
For the number of parameters and inference speeds we list both the values for just the contextual layers and all layers needed to compute context vectors.
}
\end{table*}

\section{Architectures for deep biLMs}
\label{sec:architectures}
The primary design choice when training deep biLMs for learning context vectors is the choice of the architecture for the contextual layers.
However, it is unknown if the architecture choice is important for the quality of learned representations.
To study this question, we consider two alternatives to LSTMs as described below.
See the appendix for the hyperparameter details.


\subsection{LSTM}
Among the RNN variants, LSTMs have been shown to provide state-of-the-art performance for several benchmark language modeling tasks \cite{Jzefowicz2016ExploringTL,Merity2017RegularizingAO,Melis2017OnTS}.
In particular, the LSTM with projection introduced by \citet{Sak2014LongSM} allows the model to use a large hidden state while reducing the total number of parameters.
This is the architecture adopted by \citet{Peters:2018} for computing ELMo representations.
In addition to the pre-trained 2-layer biLM from that work,\footnote{\url{http://allennlp.org/elmo}} we also trained a deeper 4-layer model to examine the impact of depth using the publicly available training code.\footnote{\url{https://github.com/allenai/bilm-tf}}
To reduce the training time for this large 4-layer model, we reduced the number of parameters in the character encoder by first projecting the character CNN filters down to the model dimension before the two highway layers.

\subsection{Transformer}
The Transformer, introduced by \citet{Vaswani2017AttentionIA}, is a feed forward self-attention based architecture.
In addition to machine translation, it has also provided strong results for Penn Treebank constituency parsing \cite{Kitaev2018TransformerParsing} and semantic role labeling \cite{Tan2017DeepSR}.
Each identical layer in the encoder first computes a multi-headed attention between a given token and all other tokens in the history, then runs a position wise feed forward network.

To adapt the Transformer for bidirectional language modeling, we modified a PyTorch based re-implementation \cite{opennmt}\footnote{\url{http://nlp.seas.harvard.edu/2018/04/03/attention.html}} to mask out future tokens for the forward language model and previous tokens for the backward language model, in a similar manner to the decoder masking in the original implementation.
We adopted hyper-parameters from the ``base'' configuration in \citet{Vaswani2017AttentionIA}, providing six layers of 512 dimensional representations for each direction.

Concurrent with our work, \citet{Radford2018ImprovingLU} trained a large forward Transformer LM and fine tuned it for a variety of NLP tasks.

\begin{table*}[!t]
\centering
\begin{tabular}{p{18ex}|>{\centering\arraybackslash}p{12ex}|>{\centering\arraybackslash}p{8ex}|>{\centering\arraybackslash}p{12ex}|c} \hline
\rule{0pt}{4ex} \textbf{Architecture} & \textbf{MultiNLI} & \textbf{SRL} & \parbox{12ex}{\textbf{Constituency \\ Parsing}} & \textbf{NER} \\[10pt] \hline\hline
GloVe &77.0 / 76.0 & 81.4 & 91.8 & 89.9 $\pm$ 0.35 \\
LSTM 2-layer& 79.6 / 79.3 & 84.6 & \textbf{93.9} & \textbf{91.7 $\pm$ 0.26} \\
LSTM 4-layer &\textbf{80.1 / 79.7}&\textbf{84.7} & \textbf{93.9} & 91.5 $\pm$ 0.12 \\
Transformer & 79.4 / 78.7 & 84.1 & 93.7 & 91.1 $\pm$ 0.26 \\
Gated CNN & 78.3 / 77.9 & 84.1 & 92.9  & 91.2 $\pm$ 0.14 \\
\hline
\end{tabular}
\caption{\label{tab:table2}
Test set performance comparison using different pre-trained biLM architectures.  The performance metric is accuracy for MultiNLI and F$_1$ score for the other tasks.  For MultiNLI, the table shows accuracy on both the matched and mismatched portions of the test set.
}
\end{table*}

\subsection{Gated CNN}
Convolutional architectures have also been shown to provide competitive results for sequence modeling including sequence-to-sequence machine translation \cite{Gehring2017ConvolutionalST}.
\citet{Dauphin2017LanguageMW} showed that architectures using Gated Linear Units (GLU) that compute hidden representations as the element wise product of a convolution and sigmoid gate provide perplexities comparable to large LSTMs on large scale language modeling tasks.

To adapt the Gated CNN for bidirectional language modeling, we closely followed the publicly available ConvSeq2Seq implementation,\footnote{\url{https://github.com/pytorch/fairseq}} modified to support causal convolutions \cite{Oord2016WaveNetAG} for both the forward and backward directions.
In order to model a wide receptive field at the top layer, we used a 16-layer deep model, where each layer is a $[4, 512]$ residual block.

\subsection{Pre-trained biLMs}
Table \ref{tab:table1} compares the biLMs used in the remainder of this study.
All models were trained on the 1 Billion Word Benchmark \cite{Chelba2014OneBW} using a sampled softmax with 8192 negative samples per batch.
Overall, the averaged forward and backward perplexities are comparable across the models with values ranging from 37.5 for the 4-layer LSTM to 44.5 for the Gated CNN.
To our knowledge, this is the first time that the Transformer has been shown to provide competitive results for language modeling.
While it is possible to reduce perplexities for all models by scaling up, our goal is to compare representations across architectures for biLMs of approximately equal skill, as measured by perplexity.

The Transformer and CNN based models are faster than the LSTM based ones for our hyperparameter choices, with speed ups of 3-5X for the contextual layers over the 2-layer LSTM model.\footnote{While the CNN and Transformer implementations are reasonably well optimized, the LSTM biLM is not as it does not use an optimized CUDA kernel due to the use of the projection cell.}
Speed ups are relatively faster in the single element batch scenario where the sequential LSTM is most disadvantaged, but are still 2.3-3X for a 64 sentence batch.
As the inference speed for the character based word embeddings could be mostly eliminated in a production setting, the table lists timings for both the contextual layers and all layers of the biLM necessary to compute context vectors.
We also note that the faster architectures will allow training to scale to large unlabeled corpora, which has been shown to improve the quality of biLM representations for syntactic tasks \cite{zhang2018lm}.


\section{Evaluation as word representations}
\label{sec:elmo_eval}
In this section, we evaluate the quality of the pre-trained biLM representations as ELMo-like contextual word vectors in state-of-the-art models across a suite of four benchmark NLP tasks.
To do so, we ran a series of controlled trials by swapping out pre-trained GloVe vectors \cite{Pennington2014GloveGV} for contextualized word vectors from each biLM computed by applying the learned weighted average ELMo pooling from \citet{Peters:2018}.\footnote{Generally speaking, we found adding pre-trained GloVe vectors in addition to the biLM representations provided a small improvement across the tasks.}
Each task model only includes one type of pre-trained word representation, either GloVe or contextual, this is a direct test of the transferability of the word representations.
In addition, to isolate the general purpose LM representations from any task specific supervision, we did not fine tune the LM weights.

Table \ref{tab:table2} shows the results.
Across all tasks, the LSTM architectures perform the best.
All architectures improve significantly over the GloVe only baseline, with relative improvements of 13\% -- 25\% for most tasks and architectures.
The gains for MultiNLI are more modest, with relative improvements over GloVe ranging from 6\% for the Gated CNN to 13\% for the 4-layer LSTM.
The remainder of this section provides a description of the individual tasks and models with details in the Appendix.

\begin{figure*}
\centering
\includegraphics[width=\textwidth]{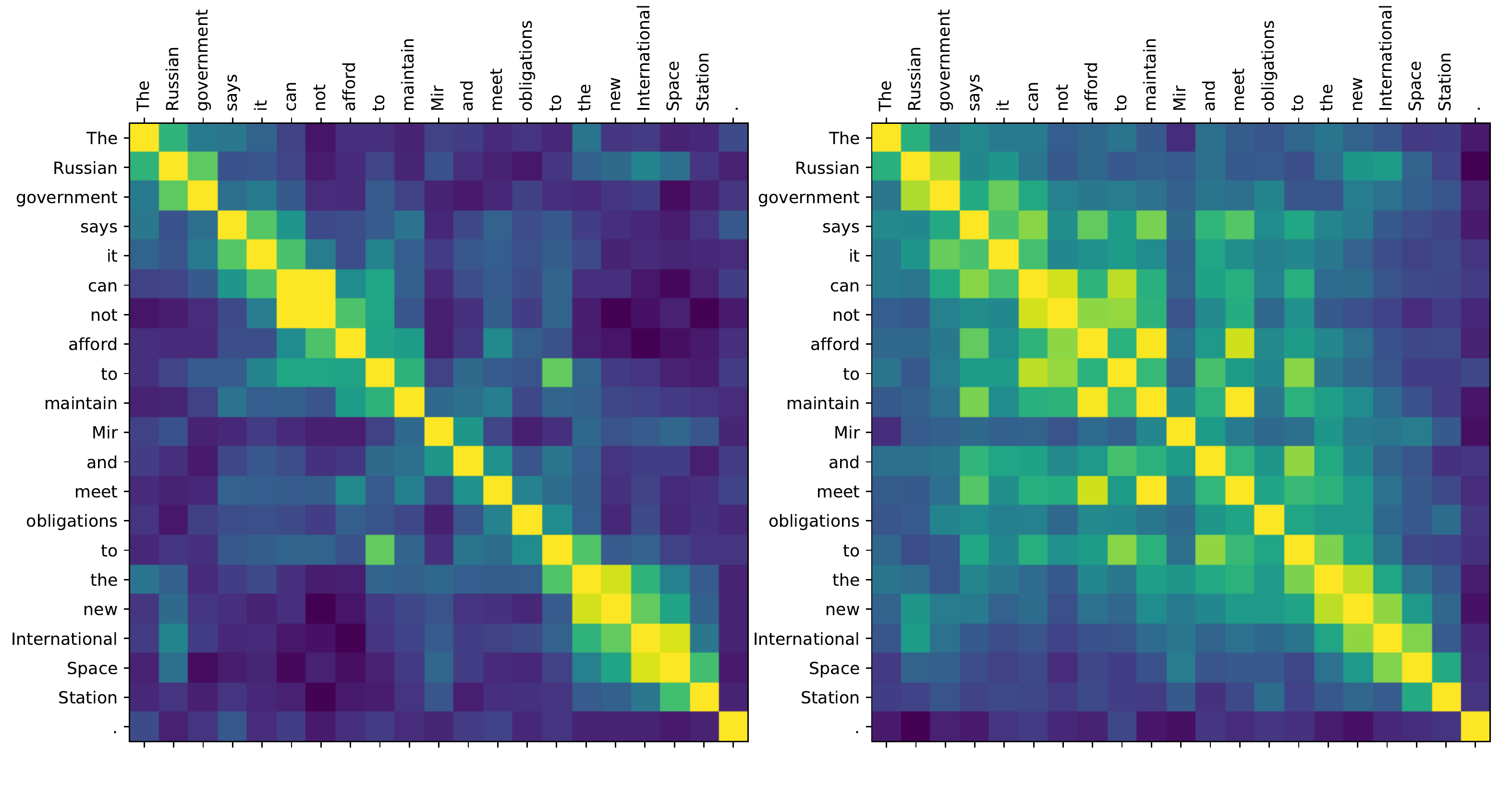}
\caption{\label{fig:contextual_similarities}
Visualization of contextual similarity between all word pairs in a single sentence using the 4-layer LSTM.
The left panel uses context vectors from the bottom LSTM layer while the right panel uses the top LSTM layer.
Lighter yellow-colored areas have higher contextual similarity.
}
\end{figure*}

\subsection{MultiNLI}  The MultiNLI dataset \cite{Williams2017ABC} contains crowd sourced textual entailment annotations across five diverse domains for training and an additional five domains for testing.
Our model is a re-implementation of the ESIM sequence model \cite{Chen2017EnhancedLF}.
It first uses a biLSTM to encode the premise and hypothesis, then computes an attention matrix followed by a local inference layer, another biLSTM inference composition layer, and finally a pooling operation before the output layer.
With the 2-layer LSTM ELMo representations, it is state-of-the-art for SNLI \cite{Peters:2018}.
As shown in Table \ref{tab:table2}, the LSTMs perform the best, with the Transformer accuracies 0.2\% / 0.6\% (matched/mismatched) less then the 2-layer LSTM.
In addition, the contextual representations reduce the matched/mismatched performance differences showing that the biLMs can help mitigate domain effects.
The ESIM model with the 4-layer LSTM ELMo-like embeddings sets a new state-of-the-art for this task, exceeding the highest previously published result by 1.3\% matched and 1.9\% mismatched from \citet{Gong2017NaturalLI}.

\subsection{Semantic Role Labeling}
The Ontonotes 5.0 Dataset \citep{Pradhan2013} contains predicate argument annotations for a variety of types of text, including conversation logs, web data, and biblical extracts.
For our model, we use the deep biLSTM from \citet{He2017DeepSR} who modeled SRL as a BIO tagging task.
With ELMo representations, it is state-of-the-art for this task \citep{Peters:2018}.
For this task, the LSTM based word representations perform the best, with absolute improvements of 0.6\% of the 4-layer LSTM over the Transformer and CNN.

\subsection{Constituency parsing}
The Penn Treebank \citep{PennTreebankMarcus1993BuildingAL} contains phrase structure annotation for approximately 40k sentences sourced from the Wall Street Journal.
Our model is the Reconciled Span Parser \citep[RSP; ][]{Joshi:2018}, which, using ELMo representations, achieved state of the art performance for this task.
As shown in Table \ref{tab:table2}, the LSTM based models demonstrate the best performance with a 0.2\% and 1.0\% improvement over the Transformer and CNN models, respectively.
Whether the explicit recurrence structure modeled with the biLSTM in the RSP is important for parsing is explored in Sec.\ \ref{sec:linear_probes}.

\subsection{Named entity recognition} The CoNLL 2003 NER task \cite{CoNLL2003NER} provides entity annotations for approximately 20K sentences from the Reuters RCV1 news corpus.
Our model is a re-implementation of the state-of-the-art system in \citet{Peters:2018} with a character based CNN word representation, two biLSTM layers and a conditional random field (CRF) loss \cite{CRF:Lafferty2001}.
For this task, the 2-layer LSTM performs the best, with averaged F$_1$ 0.4\% - 0.8\% higher then the other biLMs averaged across five random seeds.


\section{Properties of contextual vectors}
\label{sec:innate}
In this section, we examine the intrinsic properties of contextual vectors learned with biLMs, focusing on those that are independent of the architecture details.
In particular, we seek to understand how familiar types of linguistic information such as syntactic or coreferent relationships are represented throughout the depth of the network.
Our experiments show that deep biLMs learn representations that vary with network depth, from morphology in the word embedding layer, to local syntax in the lowest contextual layers, to semantic relationships such as coreference in the upper layers.

We gain intuition and motivate our analysis by first considering the inter-sentence contextual similarity of words and phrases (Sec.~\ref{sec:visualize}).
Then, we show that, in contrast to traditional word vectors, the biLM word embeddings capture little semantic information (Sec.~\ref{sec:word_analogy}) that is instead represented in the contextual layers (Sec.~\ref{sec:linear_probes}).
Our analysis moves beyond single tokens by showing that a simple span representation based on the context vectors captures elements of phrasal syntax.

\begin{figure}
\centering
\includegraphics[width=0.4\textwidth]{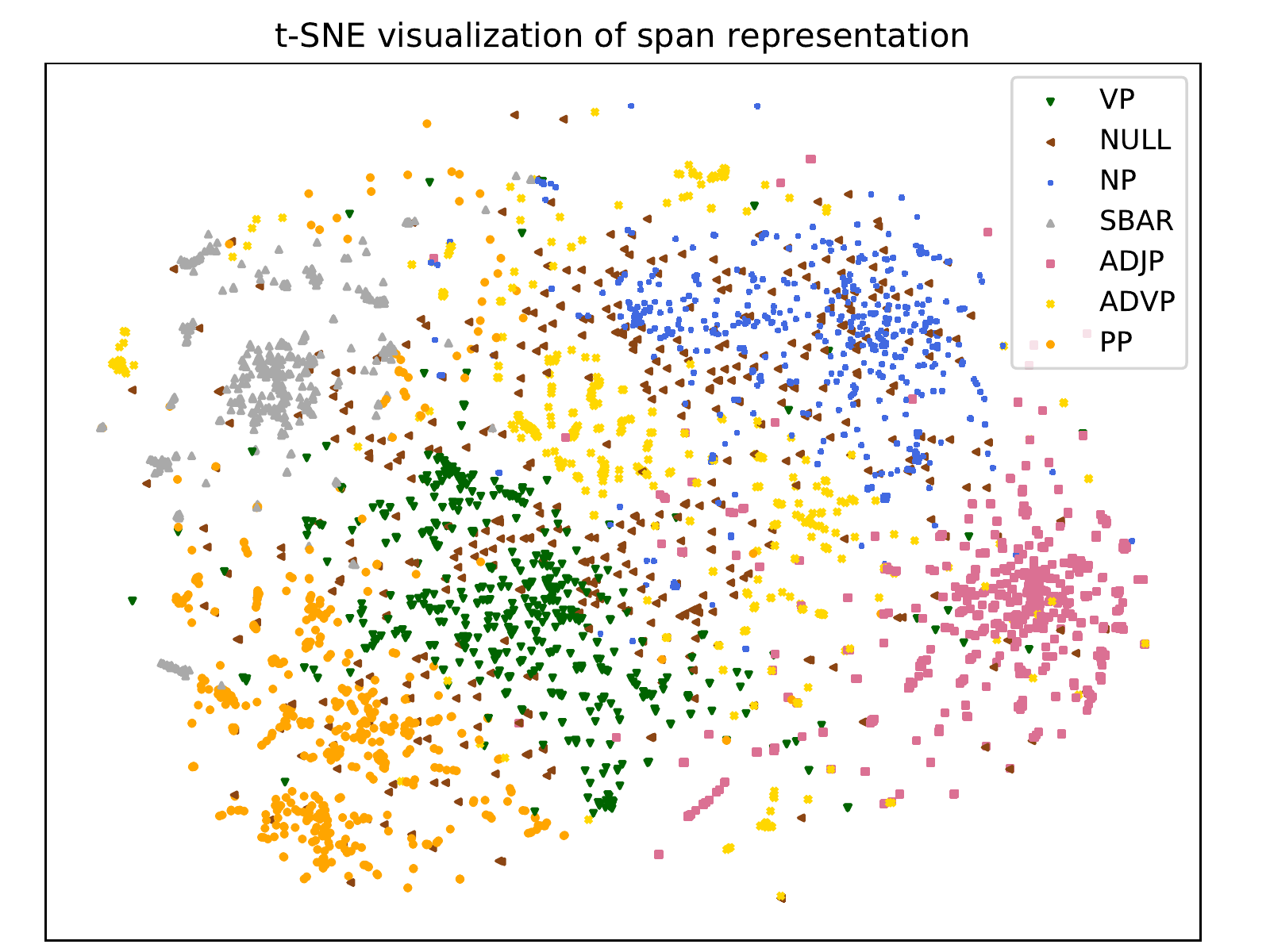}
\caption{\label{fig:span_visualization}
t-SNE visualization of 3K random chunks and 500 unlabeled spans (``NULL'') from the CoNLL 2000 chunking dataset.
}
\end{figure}

\subsection{Contextual similarity}
\label{sec:visualize}
Nearest neighbors using cosine similarity are a popular way to visualize the relationships encoded in word vectors and we can apply a similar method to context vectors.
As the biLMs use context vectors to pass information between layers in the network, this allows us to visualize how information is represented throughout the network.



\paragraph{Intra-sentence similarity}  Fig.\ \ref{fig:contextual_similarities} shows the intra-sentence contextual similarity between all pairs of words in single sentence using the 4-layer LSTM.\footnote{See appendix for visualizations of the other models.}
From the figure, we make several observations.
First, the lower layer (left) captures mostly local information, while the top layer (right) represents longer range relationships.
Second, at the lowest layer the biLM tends to place words from the same syntactic constituents in similar parts of the vector space.
For example, the words in the noun phrase ``the new international space station'' are clustered together, similar to ``can not'' and ``The Russian government''.

In addition, we can see how the biLM is implicitly learning other linguistic information in the upper layer.
For example, all of the verbs (``says'', ``can'', ``afford'', ``maintain'', ``meet'') have high similarity suggesting the biLM is capturing part-of-speech information.
We can also see some hints that the model is implicitly learning to perform coreference resolution by considering the high contextual similarity of ``it'' to ``government'', the head of ``it''s antecedent span.
Section \ref{sec:linear_probes} provides empirical support for these observations.

\paragraph{Span representations}  The observation that the biLM's context vectors abruptly change at syntactic boundaries suggests we can also use them to form representations of spans, or consecutive token sequences.
To do so, given a span of $S$ tokens from indices $s_0$ to $s_1$, we compute a span representation $\mathbf{s}_{(s_0,s_1),i}$ at layer $i$ by concatenating the first and last context vectors with the element wise product and difference of the first and last vectors:
\begin{eqnarray*}
\mathbf{s}_{(s_0,s_1),i} = [\mathbf{h}_{s_0,i}; \mathbf{h}_{s_1,i};
\mathbf{h}_{s_0,i} \odot \mathbf{h}_{s_1,i};
\mathbf{h}_{s_0,i} - \mathbf{h}_{s_1,i}
].
\end{eqnarray*}

Figure \ref{fig:span_visualization} shows a t-SNE \cite{maaten2008visualizing} visualization of span representations of 3,000 labeled chunks and 500 spans not labeled as chunks from the CoNLL 2000 chunking dataset \cite{CoNLL2000Chunking}, from the first layer of the 4-layer LSTM.
As we can see, the spans are clustered by chunk type confirming our intuition that the span representations capture elements of syntax.
Sec. \ref{sec:linear_probes} evaluates whether we can use these span representations for constituency parsing.

\begin{table}[t]
\centering
\begin{tabular}{l|cc} \hline
\textbf{Representation} & \textbf{Syntactic} & \textbf{Semantic} \\ \hline 
GloVe  & 77.9 & \textbf{79.2} \\
n-gram hash  & 72.3 & 0.5 \\ \hline
LSTM 4-layer  &  74.2 & 11.5\\
Transformer & \textbf{87.1} & 48.8  \\
Gated CNN  & 83.6 & 26.3 \\
\hline
\end{tabular}
\caption{\label{tab:word_analogy}
Accuracy (\%) for word vector analogies.  In addition to the 300 dimension 840B GloVe vectors, the table contains results from a character n-gram hash and the context insensitive word embedding layer ($\mathbf{x}_k$) from the biLMs.
}
\end{table}



\begin{figure*}[h]
\includegraphics[width=\textwidth]{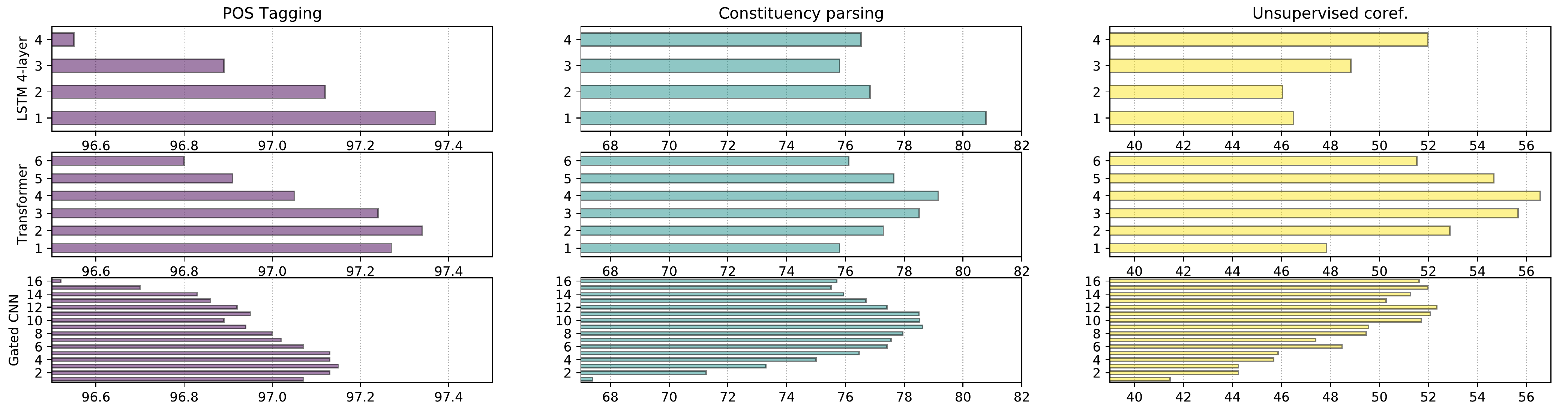}
\caption{Various methods of probing the information stored in context vectors of deep biLMs.
Each panel shows the results for all layers from a single biLM, with the first layer of contextual representations at the bottom and last layer at the top.
From top to bottom, the figure shows results from the 4-layer LSTM, the Transformer and Gated CNN models.
From left to right, the figure shows linear POS tagging accuracy (\%; Sec.\ \ref{sec:linear_probes}), linear constituency parsing (F$_1$; Sec.\ \ref{sec:linear_probes}), and unsupervised pronominal coreference accuracy (\%; Sec.\ \ref{sec:visualize}).
}
\label{fig:layer_diagnostics}
\end{figure*}

\begin{figure}[h]
\includegraphics[width=0.5\textwidth]{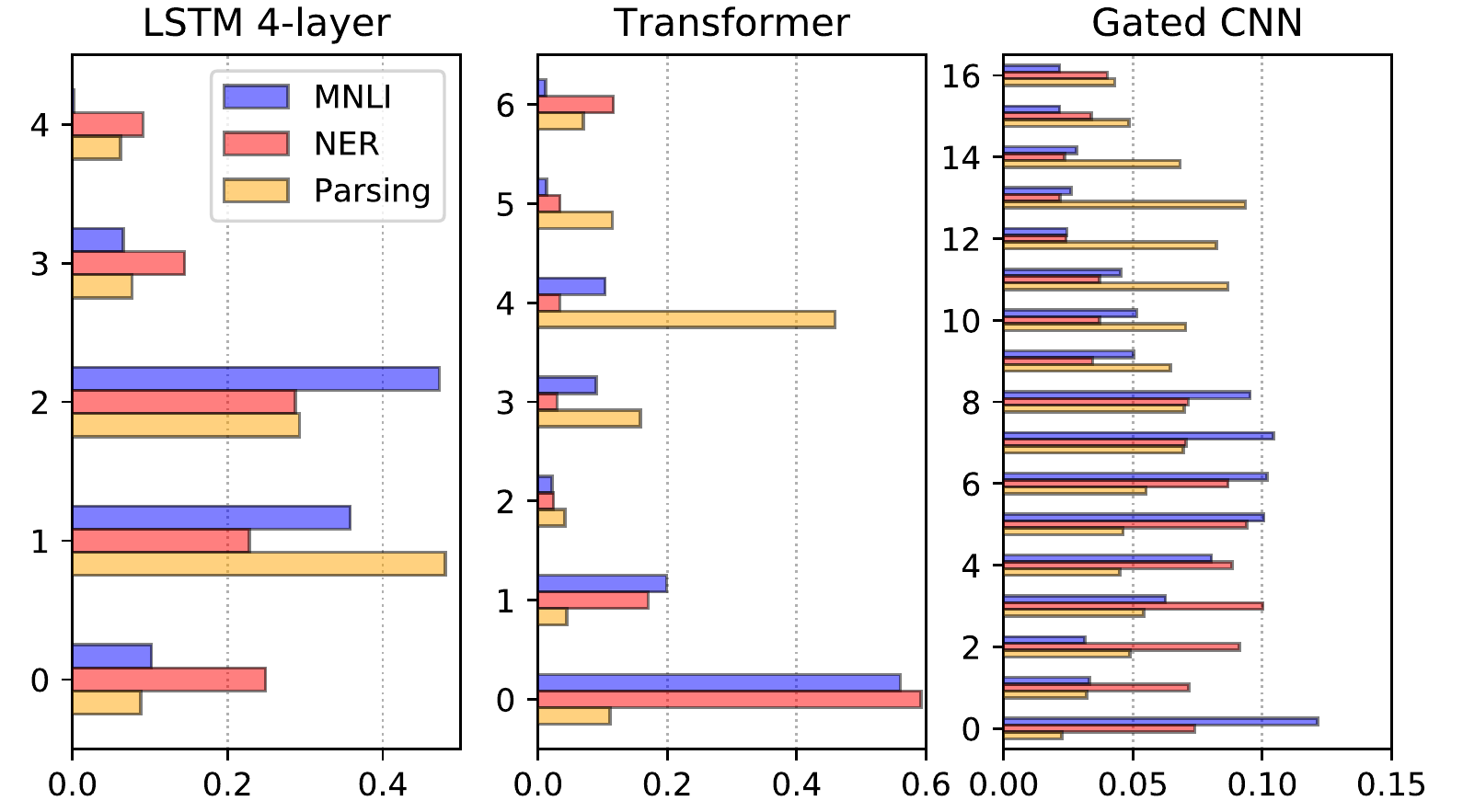}
\caption{Normalized layer weights $\mathbf{s}$ for the tasks in Sec.\ \ref{sec:elmo_eval}.
The vertical axis indexes the layer in the biLM, with layer 0 the word embedding $\mathbf{x}_k$.}
\label{fig:layer_weights}
\end{figure}

\paragraph{Unsupervised pronominal coref} \label{sec:u_coref}
We hypothesize that the contextual similarity of coreferential mentions should be similar, as in many cases it is possible to replace them with their referent.
If true, we should be able to use contextual similarity to perform unsupervised coreference resolution.
To test this, we designed an experiment as follows.
To rule out trivially high mention-mention similarities due to lexical overlap, we restricted to pronominal coreference resolution.
We took all sentences from the development set of the OntoNotes annotations in the CoNLL 2012 shared task \cite{Pradhan2012CoNLL2012ST} that had a third-person personal pronoun\footnote{he, him, she, her, it, them, they} and antecedent in the same sentence (904 sentences), and tested whether a system could identify the head word of the antecedent span given the pronoun location.
In addition, by restricting to pronouns, systems are forced to rely on context to form their representation of the pronoun, as the surface form of the pronoun is uninformative.
As an upper bound on performance, the state-of-the-art coreference model from \citet{Lee2017EndtoendNC}\footnote{\url{http://allennlp.org/models}} finds an antecedent span with the head word 64\% of the time.
As a lower bound on performance, a simple baseline that chooses the closest noun occurring before the pronoun has an accuracy of 27\%, and one that chooses the first noun in the sentence has an accuracy of 35\%.
If we add an additional rule and further restrict to antecedent nouns matching the pronoun in number, the accuracies increase to 41\% and 47\% respectively.

To use contextual representations to solve this task, we first compute the mean context vector of the smallest constituent with more then one word containing the pronoun and subtract it from the pronoun's context vector.
This step is motivated by the above observation that local syntax is the dominant signal in the contextualized word vectors, and removing it improves the accuracies of our method.
Then, we choose the noun with the highest contextual similarity to the adjusted context vector that occurs before the pronoun and matches it in number.

The right hand column of Fig.\ \ref{fig:layer_diagnostics} shows the results for all layers of the biLMs.
Accuracies for the models peak between 52\% and 57\%, well above the baseline, with the Transformer overall having the highest accuracy.
Interestingly, accuracies only drop 2-3\% compared to 12-14\% in the baseline if we remove the assumption of number agreement and simply consider all nouns, highlighting that the biLMs are to a large extent capturing number agreement across coreferent clusters.
Finally, accuracies are highest at layers near the top of each model, showing that the upper layer representations are better at capturing longer range coreferent relationships then lower layers.

\subsection{Context independent word representation}
\label{sec:word_analogy}
The word analogy task introduced in \citet{mikolov2013efficient} are commonly used as intrinsic evaluations of word vectors.
Here, we use them to compare the word embedding layer from the biLMs to word vectors.
The task has two types of analogies: syntactic with examples such as ``bird:birds :: goat:goats'', and semantic with examples such as ``Athens:Greece :: Oslo:Norway''.
Traditional word vectors score highly on both sections.
However, as shown in Table \ref{tab:word_analogy}, the word embedding layer $\mathbf{x}_k$ from the biLMs is markedly different with syntactic accuracies on par or better then GloVe, but with very low semantic accuracies.
To further highlight this distinction, we also computed a purely orthographically based word vector by hashing all character 1, 2, and 3-grams in a word into a sparse 300 dimensional vector.
As expected, vectors from this method had near zero accuracy in the semantic portion, but scored well on the syntactic portion, showing that most of these analogies can be answered with morphology alone.
As a result, we conclude that the word representation layer in deep biLMs is only faithfully encoding morphology with little semantics.

\subsection{Probing contextual information}
\label{sec:linear_probes}
In this section, we quantify some of the anecdotal observations made in Sec. \ref{sec:visualize}.
To do so, we adopt a series of linear probes \cite{Belinkov2017WhatDN} with two NLP tasks to test the contextual representations in each model layer for each biLM architecture.
In addition to examining single tokens, we also depart from previous work by examining to what extent the span representations capture phrasal syntax.

Our results show that all biLM architectures learn syntax, including span-based syntax; and part-of-speech information is captured at lower layers then constituent structure.
When combined with the coreference accuracies in Sec. \ref{sec:u_coref} that peak at even higher layers, this supports our claim that deep biLMs learn a hierarchy of contextual information.

\paragraph{POS tagging} \citet{Peters:2018} showed that the contextual vectors from the first layer of the 2-layer LSTM plus a linear classifier was near state-of-the-art for part-of-speech tagging.
Here, we test whether this result holds for the other architectures.
The second row of Fig.\ \ref{fig:layer_diagnostics} shows tagging accuracies for all layers of the biLMs evaluated with the Wall Street Journal portion of Penn Treebank \cite{PennTreebankMarcus1993BuildingAL}.
Accuracies for all of the models are high, ranging from 97.2 to 97.4, and follow a similar trend with maximum values at lower layers (bottom layer for LSTM, second layer for Transformer, and third layer for CNN).

\paragraph{Constituency parsing}
Here, we test whether the span representations introduced in Sec.\ \ref{sec:visualize} capture enough information to model constituent structure.
Our linear model is a very simple and independently predicts the constituent type for all possible spans in a sentence using a linear classifier and the span representation.
Then, a valid tree is built with a greedy decoding step that reconciles overlapping spans with an ILP, similar to \citet{Joshi:2018}.

The third row in Fig.\ \ref{fig:layer_diagnostics} shows the results.
Remarkably, predicting spans independently using the biLM representations alone has F$_1$ of near 80\% for the best layers from each model.
For comparison, a linear model using GloVe vectors performs very poorly, with F$_1$ of 18.1\%.
Across all architectures, the layers best suited for constituency parsing are at or above the layers with maximum POS accuracies as modeling phrasal syntactic structure requires a wider context then token level syntax.
Similarity, the layers most transferable to parsing are at or below the layers with maximum pronominal coreference accuracy in all models, as constituent structure tends to be more local than coreference \cite{Kuncoro2017WhatDR}.

\subsection{Learned layer weights}
Fig.\ \ref{fig:layer_weights} plots the softmax-normalized layer weights $\mathbf{s}$ from each biLM, learned as part of the tasks in Sec.\ \ref{sec:elmo_eval}.
The SRL model weights are omitted as they close to constant since we had to regularize them to stabilize training.
For constituency parsing, $\mathbf{s}$ mirrors the layer wise linear parsing results, with the largest weights near or at the same layers as maximum linear parsing.
For both NER and MultiNLI, the Transformer focuses heavily on the word embedding layer, $\mathbf{x}_k$, and the first contextual layer.
In all cases, the maximum layer weights occur below the top layers as the most transferable contextual representations tend to occur in the middle layers, while the top layers specialize for language modeling.


\section{Related work}

In addition to biLM-based representations, \citet{McCann2017LearnedIT} learned contextualized vectors with a neural machine translation system (CoVe).
However, as \citet{Peters:2018} showed the biLM based representations outperformed CoVe in all considered tasks, we focus exclusively on biLMs.

\citet{Liu2018EfficientCR} proposed using densely connected RNNs and layer pruning to speed up the use of context vectors for prediction.
As their method is applicable to other architectures, it could also be combined with our approach.

Several prior studies have examined the learned representations in RNNs.
\citet{Karpathy2015VisualizingAU} trained a character LSTM language model on source code and showed that individual neurons in the hidden state track the beginning and end of code blocks. 
\citet{Linzen2016AssessingTA} assessed whether RNNs can learn number agreement in subject-verb dependencies.  Our analysis in Sec.\ \ref{sec:u_coref} showed that biLMs also learn number agreement for coreference.
\citet{Kdr2017RepresentationOL} attributed the activation patters of RNNs to input tokens and showed that a RNN language model is strongly sensitive to tokens with syntactic functions. 
\citet{Belinkov2017WhatDN} used linear classifiers to determine whether neural machine translation systems learned morphology and POS tags.
Concurrent with our work, \citet{Khandelwal2018SharpNF} studied the role of context in influencing language model predictions, \citet{Gaddy2018WhatsGO} analyzed neural constituency parsers, \citet{blevins2018deepRNNsSyntax} explored whether RNNs trained with several different objectives can learn hierarchical syntax, and \citet{conneau2018you} examined to what extent sentence representations capture linguistic features.
Our intrinsic analysis is most similar to \citet{Belinkov2017WhatDN}; however, we probe span representations in addition to word representations, evaluate the transferability of the biLM representations to semantic tasks in addition to syntax tasks, and consider a wider variety of neural architectures in addition to RNNs.

Other work has focused on attributing network predictions.
\citet{Li2016UnderstandingNN} examined the impact of erasing portions of a network's representations on the output, \citet{Sundararajan2017AxiomaticAF} used a gradient based method to attribute predictions to inputs, and \citet{Murdoch2018BeyondWI} decomposed LSTMs to interpret classification predictions.
In contrast to these approaches, we explore the types of contextual information encoded in the biLM internal states instead of focusing on attributing this information to words in the input sentence.

\section{Conclusions and future work}
We have shown that deep biLMs learn a rich hierarchy of contextual information, both at the word and span level, and that this is captured in three disparate types of network architectures.
Across all architecture types, the lower biLM layers specialize in local syntactic relationships, allowing the higher layers to model longer range relationships such as coreference, and to specialize for the language modeling task at the top most layers.
These results highlight the rich nature of the linguistic information captured in the biLM's representations and show that biLMs act as a general purpose feature extractor for natural language, opening the way for computer vision style feature re-use and transfer methods.

Our results also suggest avenues for future work.
One open question is to what extent can the quality of biLM representations be improved by simply scaling up model size or data size?  As our results have show that computationally efficient architectures also learn high quality representations, one natural direction would be exploring the very large model and data regime.

Despite their successes biLM representations are far from perfect; during training, they have access to only surface forms of words and their order, meaning deeper linguistic phenomena must be learned ``tabula rasa".
Infusing models with explicit syntactic structure or other linguistically motivated inductive biases may overcome some of the limitations of sequential biLMs.
An alternate direction for future work combines the purely unsupervised biLM training objective with existing annotated resources in a multitask or semi-supervised manner.

\bibliography{deep_representations_bilm}
\bibliographystyle{acl_natbib_nourl}

\clearpage
\renewcommand{\appendixpagename}{\Large Appendix to accompany ``Dissecting contextual word embeddings: Architecture and Representation''}

\begin{appendices}

\section{biLM hyperparameters}
For consistency and a fair comparison, all biLMs use a 512 dimensional representation in each direction at each layer, providing a 1024 dimensional contextual representation at each layer.
All models use the same character based word embedding layer $\mathbf{x}_k$, with the exception of the 4-layer LSTM as described below.  All systems use residual connections between the contextual layers \cite{He2016DeepRL}.


\paragraph{LSTM}
Hyperparameters for the 4-layer LSTM closely follow those from the 2-layer ELMo model in \citet{Peters:2018}.
It has four layers, with each direction in each layer having a 4096 dimension hidden state and a 512 dimensional projection.
To reduce training time, the character based word representation is simplified from the other models.
It uses the same 2048 character n-gram CNN filters as the other models, but moves the projection layer from 2048 to 512 dimensions after the convolutions and before the two highway layers.

\paragraph{Transformer}
The Transformer biLM uses six layers, each with eight attention heads and a 2048 hidden dimension for the feed forward layers.
10\% dropout was used after the word embedding layer $\mathbf{x}_k$, multi-headed attention, the hidden layers in the feed forward layers and before the residual connections.
Optimization used batches of 12,000 tokens split across 4 GPUs with, using the learning rate schedule from \citet{Vaswani2017AttentionIA} with 2,000 warm up steps.
The final model weights were averaged over 10 consecutive checkpoints. 

\paragraph{Gated CNN}
The Gated CNN has 16 layers of $[4, 512]$ residual blocks with 5\% dropout between each block.
Optimization used Adagrad with linearly increasing learning rate from 0 to 0.4 over the first 10,000 batches.
The batch size was 7,500 split across 4 GPUs.
Gradients were clipped if their norm exceeded 5.0.
The final model weights were averaged over 10 consecutive checkpoints.

\section{Task model hyperparameters}

\paragraph{MultiNLI}
Our implementation of the ESIM model uses 300 dimensions for all LSTMs and all feed forward layers.
For regularization we used 50\% dropout at the input to each LSTM and after each feed forward layer.
Optimization used Adam with learning rate 0.0004 and batch size of 32 sentence pairs.

\paragraph{Semantic Role Labeling}
The SRL model uses the reimplementation of \citet{He2017DeepSR} from \citet{Gardner2017AllenNLP}.
Word representations are concatenated with a 100 dimensional binary predicate representation, specifying the location of the predicate for the given frame.
This is passed through an 8 layer bidirectional LSTM, where the layers alternate between forward and backward directions. Highway connections and variational dropout are used between every LSTM layer.
Models are with a batch size of 80 sentences using Adadelta \citep{Zeiler2012adadelta} with an initial learning rate of 1.0 and rho 0.95.

\paragraph{Constituency Parsing}
The constituency Parser is a reimplementation of \citet{Joshi:2018}, available in AllenNLP \citep{Gardner2017AllenNLP}.
Word representations from the various biLMs models are passed through a two layer bidirectional LSTM with hidden size 250 and 0.2 dropout.
Then, the span representations are passed through a feedforward layer (dropout 0.1, hidden size 250) with a relu non-linearity before classification.
We use a batch size of 64 and gradients are normalized to have a global norm $\leq$ 5.0.
Optmization uses Adadelta with initial learning rate of 1.0 and rho 0.95.

\paragraph{NER}
The NER model concatenates 128 character CNN filters of width 3 characters to the pre-trained word representations.
It uses two LSTM layers with hidden size 200 with 50\% dropout at input and output.
The final layer is a CRF, with constrained decoding to enforce valid tag sequences.
We employ early stopping on the development set and report average F$_1$ across five random seeds.

\section{Contextual similarities}
\begin{figure*}[b!]
\includegraphics[width=\textwidth]{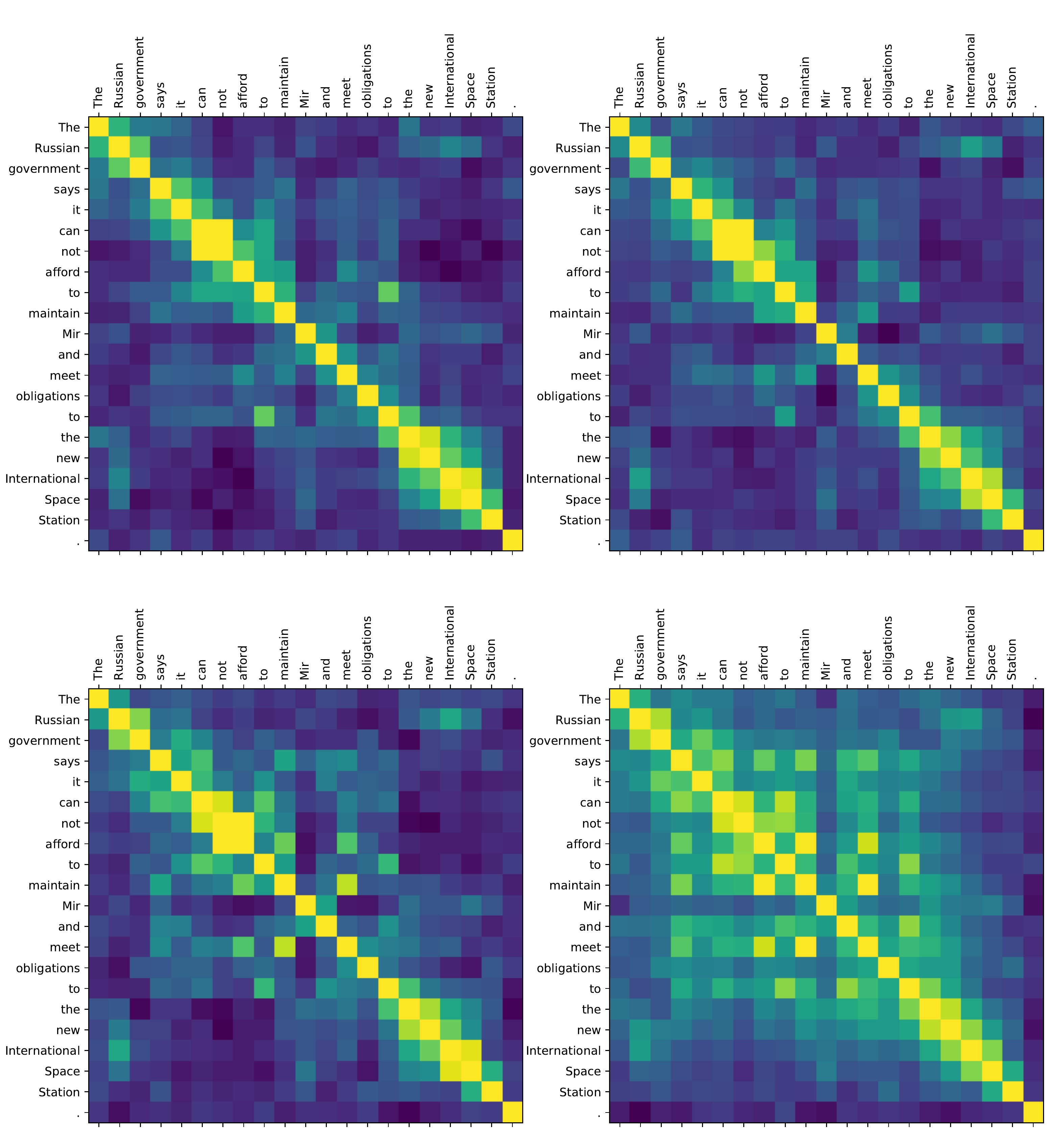}
\caption{Visualization of contextual similarities from the 4-layer LSTM biLM.
The first layer is at top left and last layer at bottom right, with the layer indices increasing from left to right and top to bottom in the image.
\label{fig:elmo_all_layers}
}
\end{figure*}

\begin{figure*}
\includegraphics[width=\textwidth]{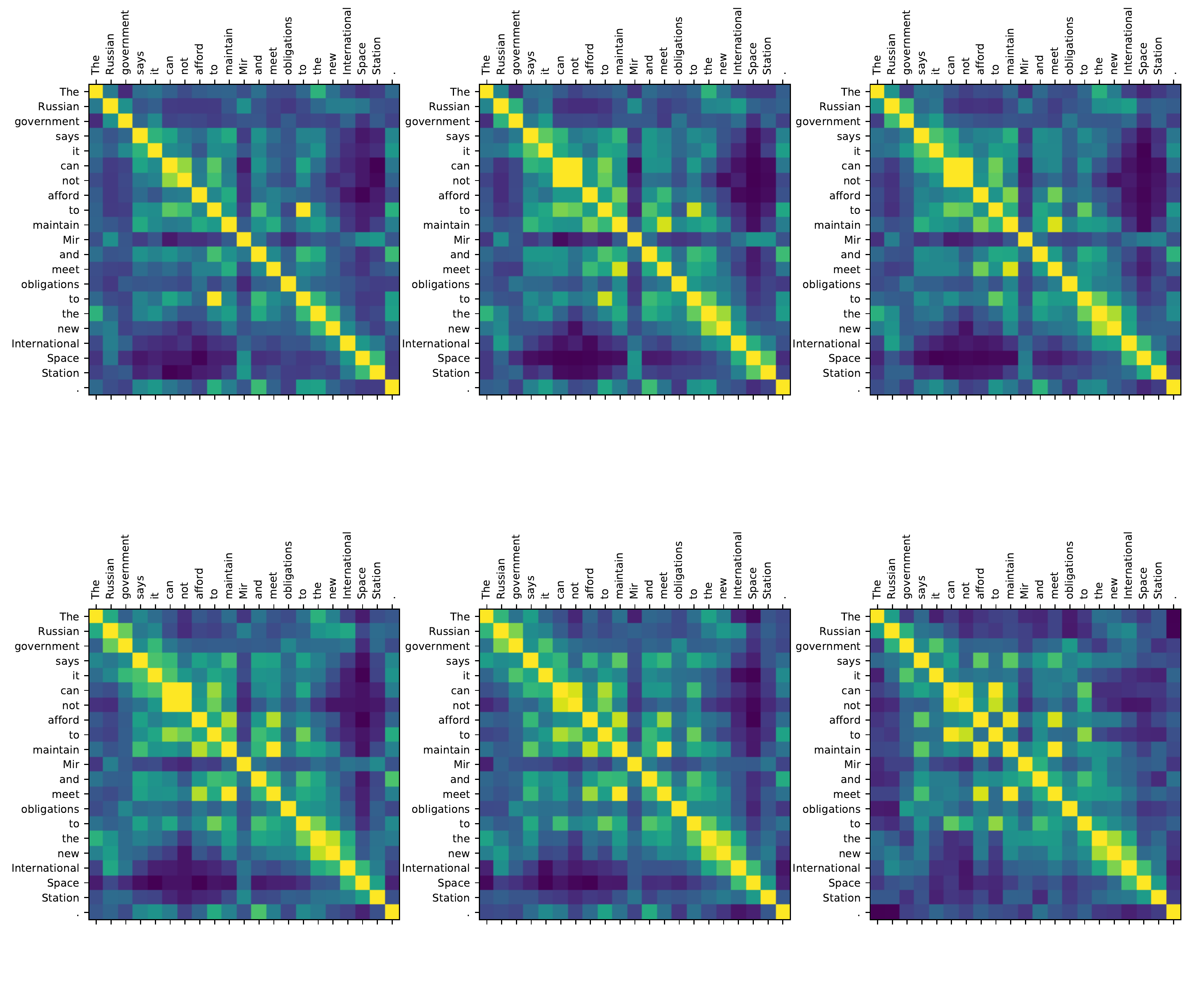}
\caption{Visualization of contextual similarities from the Transformer biLM.
The first layer is at top left and last layer at bottom right, with the layer indices increasing from left to right and top to bottom in the image.
\label{fig:transformer_all_layers}
}
\end{figure*}

\begin{figure*}
\includegraphics[width=\textwidth]{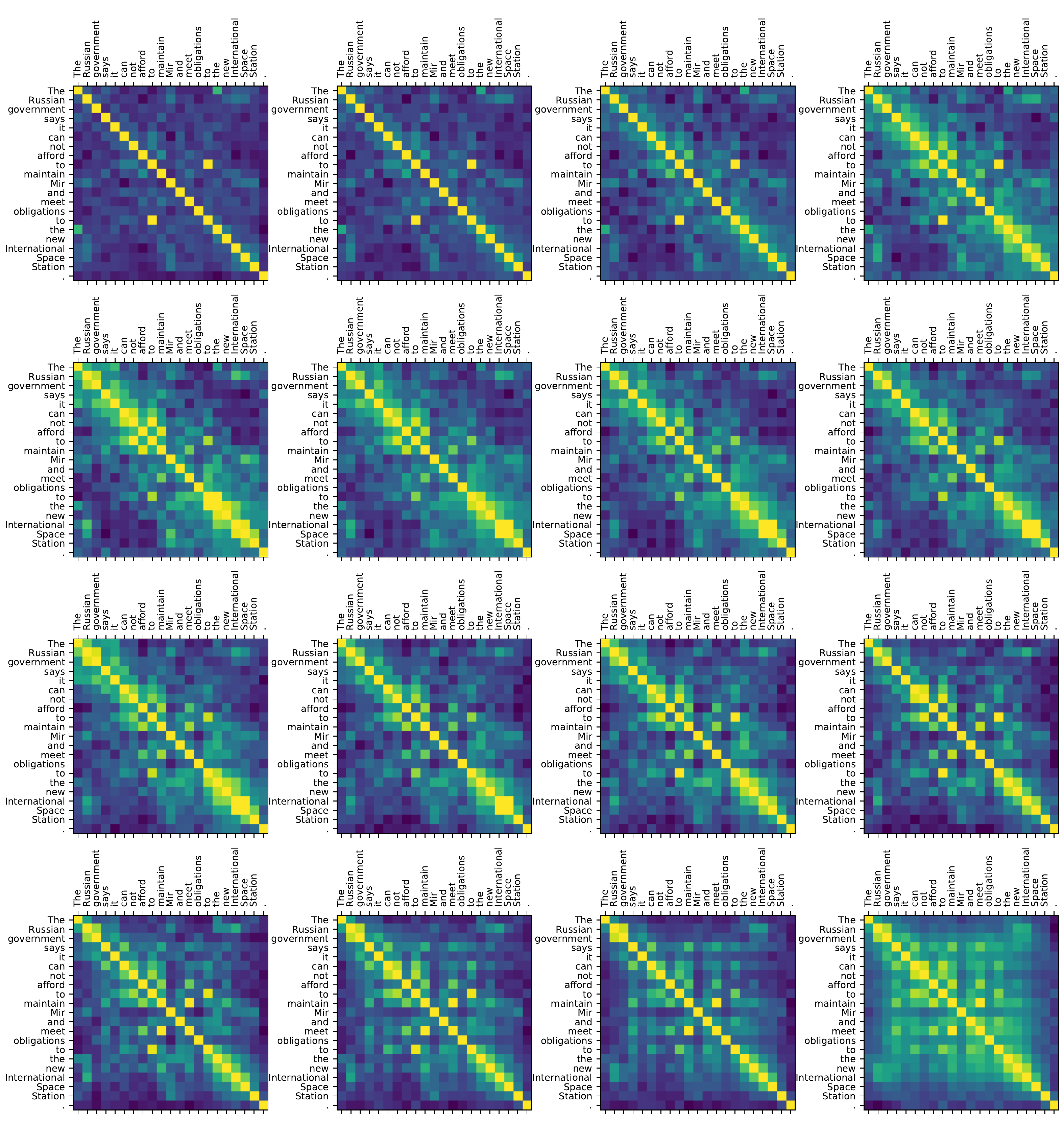}
\caption{Visualization of contextual similarities from the gated CNN biLM.
The first layer is at top left and last layer at bottom right, with the layer indices increasing from left to right and top to bottom in the image.
\label{fig:cnn_all_layers}
}
\end{figure*}

Figures \ref{fig:elmo_all_layers}, \ref{fig:transformer_all_layers}, and \ref{fig:cnn_all_layers} show contextual similarities similar to Figure \ref{fig:contextual_similarities} for all layers from the 4-layer LSTM, the Transformer and gated CNN biLMs.

\section{Layer diagnostics}
Tables \ref{table:coref-full}, \ref{table:pos-full} and \ref{table:parsing-full} list full results corresponding to the top three rows in Fig.\ \ref{fig:layer_diagnostics}.

\clearpage


\begin{table}
\centering
\begin{tabular}{@{}lc@{}}
\toprule
\textbf{Layer}          & \textbf{Accuracy} \\ \midrule
\textbf{Elmo - 4 Layer} &                   \\
Layer 1                 & 46.5            \\
Layer 2                 & 46.0            \\
Layer 3                 & 48.8            \\
Layer 4                 & \textbf{52.0}            \\ \midrule
\textbf{Transformer}    &                   \\
Layer 1                 & 47.8            \\
Layer 2                 & 52.9            \\
Layer 3                 & 55.7            \\
Layer 4                 & \textbf{56.7}            \\
Layer 5                 & 54.7            \\
Layer 6                 & 51.5            \\       \midrule
\textbf{Gated CNN}      &                   \\
Layer 1                 & 41.5            \\
Layer 2                 & 44.2           \\
Layer 3                 & 44.2            \\
Layer 4                 & 45.7            \\
Layer 5                 & 45.9            \\
Layer 6                 & 48.5            \\
Layer 7                 & 47.4            \\
Layer 8                 & 49.6           \\
Layer 9                 & 51.7            \\
Layer 10                & 47.8            \\
Layer 11                & 52.1            \\
Layer 12                & \textbf{52.4}            \\
Layer 13                & 50.3            \\
Layer 14                & 51.3            \\
Layer 15                & 52.0            \\
Layer 16                & 51.6            \\ \bottomrule
\end{tabular}
\caption{Unsupervised pronominal accuracies using the CoNLL 2012 development set.}
\label{table:coref-full}
\end{table}

\begin{table}
\centering
\begin{tabular}{@{}lc@{}}
\toprule
\textbf{Layer}          & \textbf{Accuracy} \\ \midrule
\textbf{GloVe Only}     & 88.61              \\ \midrule
\textbf{Elmo - 4 Layer} &                   \\
Layer 1                 & \textbf{97.36}            \\
Layer 2                 & 97.16            \\
Layer 3                 & 96.90            \\
Layer 4                 & 96.58            \\
Weighted Layers         & 97.22            \\ \midrule
\textbf{Transformer}    &                   \\
Layer 1                 & 97.30            \\
Layer 2                 & 97.35            \\
Layer 3                 & 97.25            \\
Layer 4                 & 97.15            \\
Layer 5                 & 96.90            \\
Layer 6                 & 96.82            \\
Weighted Layers         & \textbf{97.48}            \\ \midrule
\textbf{Gated CNN}      &                   \\
Layer 1                 & 97.09            \\
Layer 2                 & 97.16            \\
Layer 3                 & 97.19            \\
Layer 4                 & 97.16            \\
Layer 5                 & 97.11            \\
Layer 6                 & 97.09            \\
Layer 7                 & 97.08            \\
Layer 8                 & 97.01            \\
Layer 9                 & 97.00            \\
Layer 10                & 96.97            \\
Layer 11                & 96.96            \\
Layer 12                & 96.97            \\
Layer 13                & 96.85            \\
Layer 14                & 96.80            \\
Layer 15                & 96.64            \\
Layer 16                & 96.43            \\
Weighted Layers         & \textbf{97.26}            \\ \bottomrule
\end{tabular}
\caption{POS tagging accuracies for the linear models on the PTB dev set.}
\label{table:pos-full}
\end{table}

\begin{table}
\centering
\begin{tabular}{@{}lccc@{}}
\toprule
\textbf{Layer}          & \textbf{F$_1$}    & \textbf{Precision} & \textbf{Recall} \\ \midrule
\textbf{GloVe Only}     & 18.1          & 11.2               & 45.9            \\ \midrule
\textbf{LSTM - 4 Layer} &                &                    &                 \\ 
Layer 1                 & 80.8          & 87.0              & 74.1           \\
Layer 2                 & 76.8          & 83.7              & 71.4           \\
Layer 3                 & 75.8          & 84.6              & 68.7           \\
Layer 4                 & 76.5          & 81.6              & 72.1           \\
Weighted Layers         & \textbf{80.9} & \textbf{87.9}     & \textbf{75.4}  \\
\midrule
\textbf{Transformer}    &                &                    &                 \\
Layer 1                 & 75.8          & 80.4              & 71.7           \\
Layer 2                 & 77.3          & 82.6              & 72.6           \\
Layer 3                 & 78.5          & 82.5              & 75.0           \\
Layer 4                 & 79.2          & 80.5              & 77.9           \\
Layer 5                 & 77.7          & 78.3              & 77.0           \\
Layer 6                 & 76.1          & 77.3              & 75.0           \\
Weighted Layers         & \textbf{82.8} & \textbf{87.5}     & \textbf{78.6}  \\
\midrule
\textbf{Gated CNN}            &                &                    &                 \\
Layer 1                 & 67.4          & 79.5              & 58.5           \\
Layer 2                 & 71.3          & 81.8              & 63.1           \\
Layer 3                 & 73.3          & 83.8              & 65.1           \\
Layer 4                 & 75.0          & 84.6              & 67.3           \\
Layer 5                 & 76.5          & 85.3              & 69.3           \\
Layer 6                 & 77.4          & 85.6              & 70.7           \\
Layer 7                 & 77.6          & 85.9              & 70.7           \\
Layer 8                 & 78.0          & 86.9              & 71.1           \\
Layer 9                 & 78.6          & 85.0              & \textbf{73.3}           \\
Layer 10                & 78.5          & 85.9              & 72.3  \\
Layer 11                & 78.5           & 84.5              & \textbf{73.3}            \\
Layer 12                & 77.4          & 85.4              & 70.8           \\
Layer 13                & 76.7           & 84.7              & 70.1            \\
Layer 14                & 75.9          & 83.3              & 69.9            \\
Layer 15                & 75.5          & 82.8              & 69.4           \\
Layer 16                & 75.7          & 83.0              & 69.6           \\
Weighted Layers         & \textbf{78.6} & \textbf{86.4}     & 72.0           \\ \midrule
\end{tabular}
\caption{Labeled Bracketing F$_1$, Precision and Recall for the linear parsing models on the PTB dev set.}
\label{table:parsing-full}
\end{table}

\end{appendices}

\end{document}